\documentclass[journal]{IEEEtran}

\usepackage{amsmath,amsfonts}
\usepackage{array}
\usepackage{booktabs}
\usepackage{cite}
\usepackage{graphicx}
\usepackage[section]{placeins}
\usepackage{textcomp}
\usepackage{tikz}
\usepackage{url}

\newif\ifincludeauthorbios
\includeauthorbiosfalse

\begin{document}

\title{WM-Cov: Test Adequacy for Interactive World-Model-Style Autonomous Driving Simulation}

\author{Jianxun~Cui, Ping~Wu, Stanisa~Peric, Marko~Milojkovic, and Vladan~Devedzic%
\thanks{J. Cui is with the School of Transportation Science and Engineering,
Harbin Institute of Technology, Harbin, China, and the Chongqing Research
Institute of HIT, Chongqing, China (e-mail: cuijianxun@hit.edu.cn).
Corresponding author: J. Cui.}%
\thanks{P. Wu is with Chongqing Changan Automobile Co., Ltd., Chongqing, China
(e-mail: wuping@changan.com.cn).}%
\thanks{S. Peric and M. Milojkovic are with the University of Nis, Faculty of
Electronic Engineering, Department of Control Systems, Nis, Serbia (e-mails:
stanisa.peric@elfak.ni.ac.rs; marko.milojkovic@elfak.ni.ac.rs).}%
\thanks{V. Devedzic is with the University of Belgrade, Faculty of
Organizational Sciences, Belgrade, Serbia (e-mail:
vladan.devedzic@fon.bg.ac.rs).}}

\markboth{IEEE Transactions on Intelligent Vehicles}%
{WM-Cov: Test Adequacy for Interactive WM-Style Simulation}

\maketitle

\begin{abstract}
World models and generative simulators are emerging as interactive testing infrastructure for autonomous driving because they can react to the ego planner and produce counterfactual, rare, and safety-critical rollouts. This changes the meaning of a test scenario. A scenario is no longer only a fixed replayed trajectory, but an interactive scenario family: a map, route, initial condition, environment condition, background-agent policy, prompt, and seed distribution whose realized evolution depends on the ego planner under test. The unresolved testing question is therefore not only whether dangerous rollouts can be generated, but what valid closed-loop evidence is enough to support a specified testing intent and stopping decision. This paper formulates interactive WM-style testing adequacy and introduces WM-Cov, a provider-agnostic evaluation layer that turns raw provider outputs into requested, realized, and valid realized evidence. WM-Cov reports stopping-oriented adequacy through coverage growth, valid-failure discovery, failure-mode diversity, realism, artifact suppression, duplicate accounting, and valid-evidence precision. Tiered evaluation studies on executed TeraSim/SUMO events and WM-like mixed trace pools show that dangerous-looking generated events can contain both valid ADS failures and non-evidential artifacts. A real DriveArena TrafficManager--WorldDreamer matrix further evaluates two planners, two horizons, six prompt conditions, and 360 total ego-route requests; a disjoint 80-request route-slice check evaluates whether the same accounting behavior persists beyond the first ego-route slice. In the main matrix, 304 attempts become fully realized evidence and 56 remain partial; in the disjoint slice, 74 of 80 additional attempts are fully realized and 6 remain partial. The results support evaluating world-model-style testing by convergence of valid interactive evidence under budget, rather than by raw generated failures or prompt coverage alone.
\end{abstract}


\section{Introduction}
\label{sec:introduction}

Autonomous driving testing has traditionally relied on scenario libraries, scenario taxonomies, and coverage-oriented evaluation protocols. These mechanisms provide a way to reason about what has been tested, what remains untested, and when additional testing may have diminishing value. This coverage-and-sufficiency view is one of the main strengths of the scenario-library paradigm.

World models and generative simulators change the testing substrate more fundamentally than simply adding another source of scenes. In an interactive world-model-style simulator, the ego planner under test influences the future world: the background agents, sensor observations, and subsequent risk may respond differently when a different planner is inserted. A scenario is therefore not only a fixed replayed trajectory. It is an interactive scenario family specified by a testing intent, map, route, initial state, environment condition, background-agent policy, prompt, and seed distribution. Each closed-loop run is one realized interaction trace from this family.

This paper studies the resulting adequacy problem. Given a testing intent such as unprotected intersection crossing, how much closed-loop world-model evidence is needed before the tester can justify stopping or extending the campaign for that planner and scenario family? The answer cannot be a raw number of generated videos, prompts, or collision events. It must depend on whether requested conditions are realized, whether realized interactions cover relevant behaviors, whether newly generated runs still reveal new valid evidence, and whether apparent failures are artifacts or duplicates.

We introduce WM-Cov as an evidence-accounting and stopping-oriented evaluation layer for this setting. WM-Cov separates three quantities: requested coverage, which records the test conditions asked of the provider; realized interaction coverage, which records what actually happened in closed loop; and valid realized coverage, which only counts plausible, non-duplicate, non-artifact evidence. The goal is not to propose a new world model or dangerous-scenario generator. The goal is to define when interactive WM-style testing is becoming adequate under a fixed budget.

The contributions are:
\begin{itemize}
    \item We formulate interactive world-model-style autonomous-driving testing as an adequacy and stopping problem over scenario families rather than fixed replayed trajectories.
    \item We introduce WM-Cov, a provider-agnostic evaluation layer that tracks requested, realized, and valid realized coverage for closed-loop interaction traces.
    \item We define an adequacy report that combines coverage growth, valid failures, failure diversity, artifact counts, duplicate accounting, risk, realism, and precision.
    \item We evaluate the framework across executed logs, mixed trace pools, DriveArena traces, and a real closed-loop DriveArena--WorldDreamer provider matrix with UniAD and VAD.
\end{itemize}

\section{Related Work}
\label{sec:related_work}

Scenario-based testing has become a central paradigm for autonomous-driving safety assessment. Existing work studies how driving situations can be represented as scenario classes, parameter spaces, operational-design-domain attributes, behavior competencies, and executable simulator configurations. This line of work also provides the vocabulary of coverage, completeness, criticality, diversity, and test adequacy that is needed to decide whether a scenario database or simulation campaign is sufficiently informative \cite{tang2023survey,laurent2023parameter,neelofar2024towards,degelder2024coverage,degeldersingh2025scenario,chodowiec2026odd}.

Recent scenario-metric and ODD/behavior-coverage work is especially relevant because it treats coverage as a safety-assurance object rather than as a raw scenario count. Scenario metrics are now discussed across completeness, coverage, criticality, diversity, exposure, and complexity within safety-assurance frameworks \cite{degeldersingh2025scenario}; ODD- and behavior-based coverage further emphasizes that a test suite should cover operational attributes, behavior competencies, out-of-ODD handling, and rule compliance \cite{chodowiec2026odd}. Iterative scenario-based V\&V work on triggering conditions also shows that test results can feed subsequent scenario generation in a SOTIF-oriented process \cite{zhu2025triggering}. These works are conceptual predecessors of WM-Cov and establish coverage-based adequacy as an important concern for autonomous-driving simulation.

The limitation addressed here is narrower and WM-specific. Most coverage-oriented work assumes a curated scenario set, a scenario database, or a parameterized simulator campaign whose generated executions are treated as ordinary simulator tests. A modern world-model-style provider changes the source of tests: it can generate an open-ended trace pool whose realized interactions depend on the ego planner and whose failures may be valid ADS evidence, duplicates, conditioning failures, implausible cases, or generator artifacts. WM-Cov reuses the coverage concern from scenario-based testing, but places it after interactive generative production and before evidence is counted.

Safety-critical scenario generation methods expose rare or dangerous situations more efficiently than unguided simulation through search, optimization, reinforcement learning, importance sampling, and data-driven generation \cite{ding2023survey}. Recent generative simulators such as TeraSim emphasize discovering unknown unsafe events through generative simulation \cite{sun2025terasim}. WM-Cov addresses a different question: once such events are generated, which count as valid evidence, how much of the test space do they cover, and which subset should be selected under a fixed test budget?

Recent autonomous-driving world models and generative simulation platforms can synthesize realistic and controllable driving scenes, videos, trajectories, and closed-loop interactions. Representative systems include generative world models such as GAIA-1 \cite{hu2023gaia}, closed-loop generative simulation platforms such as DriveArena \cite{yang2025drivearena}, and reliability-oriented driving world models such as ReSim \cite{yang2026resim}. Surveys of world models, foundation models, and generative AI for autonomous-driving testing also show that scenario generation and scenario analysis are becoming central applications of these models \cite{guan2025world,gao2026foundation,song2026generative}.

The phrase ``world model'' also has a longer testing history outside modern neural world models. Earlier model-based testing work uses engineered world models, CEFSMs, and Petri nets to generate tests for autonomous systems \cite{andrews2015active,andrews2016world}. This line confirms that world-state modeling has long been useful for autonomous-system testing. However, it assumes explicit formal or semi-formal models as test-generation artifacts, rather than neural or generative providers that may produce realistic but partly invalid closed-loop evidence.

Synthetic-scenario studies further show that the realism and construction process of simulated tests matter \cite{tuncali2025synthetic}. These works, together with scenario metrics and world-model-based test generation, are closest to the adequacy concern in this paper. The remaining gap is the evaluation layer for an interactive generative or neural world-model-style trace source: the trace pool is large, adaptive, planner-dependent, and partly unreliable, so coverage must be coupled with requested-to-realized tracking, validity accounting, duplicate suppression, and budgeted accumulation before generated events are counted as evidence.

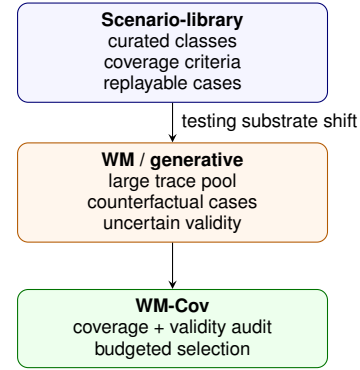
\begin{figure}[t]
    \centering
    \begin{tikzpicture}[
        font=\sffamily\scriptsize,
        box/.style={draw, rounded corners, minimum height=13mm, minimum width=41mm, align=center},
        note/.style={draw, rounded corners, minimum height=10mm, minimum width=41mm, align=center},
        >=stealth
    ]
    \node[box, fill=blue!4, draw=blue!45!black] (lib) at (0,0)
    {\textbf{Scenario-library}\\curated classes\\coverage criteria\\replayable cases};
    \node[box, fill=orange!7, draw=orange!70!black] (wm) at (0,-1.85)
    {\textbf{WM / generative}\\large trace pool\\counterfactual cases\\uncertain validity};
    \node[note, fill=green!7, draw=green!45!black] (cov) at (0,-3.65)
    {\textbf{WM-Cov}\\coverage + validity audit\\budgeted selection};
    \draw[->] (lib) -- node[right]{testing substrate shift} (wm);
    \draw[->] (wm) -- (cov);
    \end{tikzpicture}
    \caption{Scenario-library testing often starts from curated replayable cases and coverage criteria, whereas interactive WM-style testing starts from scenario families whose closed-loop realizations depend on the ego planner. Adequacy must therefore be audited over valid interaction evidence rather than raw generated events.}
    \label{fig:paradigm_gap}
\end{figure}

Most world-model and generative-simulation literature evaluates the generator through visual realism, temporal coherence, controllability, trajectory plausibility, closed-loop behavior, or safety-critical exposure. These metrics are necessary but not the same as testing adequacy: a realistic generator can still have poor feature coverage, and a failure-rich generator can concentrate evidence in one narrow mode. WM-Cov is positioned as a provider-agnostic layer above such models. In short, existing coverage work explains why testing campaigns need adequacy criteria, while existing world-model platforms explain how interactive generative traces can be produced; WM-Cov connects the two by defining when requested traces become realized, valid, and budget-relevant testing evidence. Figure~\ref{fig:paradigm_gap} summarizes this shift from curated scenario-library testing to audited WM-style evidence accumulation.

\section{Problem Formulation}
\label{sec:problem}

We consider a testing workflow in which a world-model-style provider is used as an interactive test environment. A testing intent $q$ specifies what the tester wants to evaluate, such as intersection crossing, unprotected left turn, lane merge, or occluded pedestrian encounter. Instead of treating a scenario as a fixed replay, WM-Cov treats it as an interactive scenario family
\begin{equation}
    \mathcal{F}(q)=\{m,\rho,x_0,e,\pi_{bg},p,\mathcal{Z}\},
\end{equation}
where $m$ is the map or local road structure, $\rho$ is the route or goal, $x_0$ is the initial state, $e$ is the environment condition, $\pi_{bg}$ is the background-agent response policy, $p$ is a prompt or conditioning description, and $\mathcal{Z}$ is the seed or latent perturbation distribution. Let $\mathcal{G}$ denote the world-model-style provider, which may be a neural world model, generative simulator, or closed-loop simulation service. For a fixed ego planner $\pi_e$, one closed-loop execution samples $z_i\sim\mathcal{Z}$ and produces an interaction trace
\begin{equation}
    \tau_i = \mathrm{Rollout}(\mathcal{G}, \mathcal{F}(q), \pi_e, z_i).
\end{equation}
The testing adequacy question is how many traces $\tau_i$ are needed before the accumulated evidence supports a stopping claim for $\pi_e$ under the stated family and intent.

Each trace is mapped into a feature vector
\begin{equation}
    \phi(\tau_i) = (o_i, u_i, r_i, b_i, f_i, v_i),
\end{equation}
where $o_i$ is the operational design domain, $u_i$ is the interaction type, $r_i$ is a risk bin, $b_i$ is the behavior class, $f_i$ is the failure mode, and $v_i$ is the validity label.

The validity label is essential. We distinguish valid safe cases, valid safety-critical cases, valid ADS failures, duplicate failures, invalid scenarios, and artifact failures. This prevents a generated collision or emergency stop from being counted as safety evidence unless the event is plausible, replayable, and relevant to the ADS under test. In an interactive setting, a failure can also be planner-specific: a trace may expose a weakness of $\pi_e$ without implying that every planner will fail under the same family. WM-Cov therefore counts evidence relative to the planner and scenario family being tested.

WM-Cov separates requested, realized, and valid realized coverage. Let $\Omega_{req}(q)$ be the target bins implied by the testing intent and conditioning design. The provider may fail to realize some requested conditions: for example, a prompt requesting a rainy night pedestrian crossing may produce a daytime vehicle-only interaction. Let $\Omega_{real}(\mathcal{T})$ be the bins actually observed in a trace set $\mathcal{T}$, and $\Omega_{valid}(\mathcal{T})$ be the subset supported by non-artifact, non-duplicate evidence. The three coverage levels are
\begin{equation}
    C_{req\rightarrow real} =
    \frac{|\Omega_{real}(\mathcal{T})\cap\Omega_{req}(q)|}{|\Omega_{req}(q)|},
\end{equation}
\begin{equation}
    C_{real\rightarrow valid} =
    \frac{|\Omega_{valid}(\mathcal{T})|}{|\Omega_{real}(\mathcal{T})|},
\end{equation}
and
\begin{equation}
    C_{valid} =
    \frac{|\Omega_{valid}(\mathcal{T})\cap\Omega_{req}(q)|}{|\Omega_{req}(q)|}.
\end{equation}

Let $R(\tau_i)\in[0,1]$ be a risk score and $Q(\tau_i)\in[0,1]$ be a realism or validity-confidence score. Let $A(\tau_i)\in[0,1]$ denote artifact risk, where invalid scenarios and artifact failures receive high penalty. For a non-empty trace set $\mathcal{T}$, $\overline{R}(\mathcal{T})$, $\overline{Q}(\mathcal{T})$, and $\overline{A}(\mathcal{T})$ denote the arithmetic means of $R(\tau)$, $Q(\tau)$, and $A(\tau)$ over traces $\tau\in\mathcal{T}$. Given a test budget $B$, the operational selection problem is to choose traces that maximize valid, diverse, and non-artifactual evidence:
\begin{equation}
    \mathcal{T}^{*} = \arg\max_{\mathcal{T}, |\mathcal{T}|\leq B}
    \; C_{valid}(\mathcal{T}) + \lambda_r \overline{R}(\mathcal{T})
    + \lambda_q \overline{Q}(\mathcal{T})
    - \lambda_a \overline{A}(\mathcal{T}).
\end{equation}
Here $\lambda_r$, $\lambda_q$, and $\lambda_a$ are user-specified weights that control the relative importance of risk exposure, realism or validity confidence, and artifact suppression.
The formulation is intentionally provider-agnostic: $\mathcal{G}$ can be a neural world model, a closed-loop generative simulator, or a rule-based traffic simulator, as long as its outputs can be mapped into the WM-Cov schema.

\subsection{Stopping-Oriented Adequacy}

WM-Cov treats adequacy as a convergence claim, not as a fixed run count. Let $\mathcal{T}_n$ denote the first $n$ validly processed traces. A family-level test campaign may stop when coverage growth, valid-failure discovery, and uncertainty become small:
\begin{align}
    \Delta C_{valid}(n,w) &< \epsilon_c, \\
    \Delta F_{valid}(n,w) &< \epsilon_f, \\
    \mathrm{width}(\widehat{p}_{fail}(n)) &< \epsilon_p,
\end{align}
over a recent window of $w$ traces, while artifact rate remains below a threshold and required bins have no unresolved gaps. Here $\Delta F_{valid}$ is the number of newly discovered valid failure modes in the window, $\widehat{p}_{fail}$ is the estimated valid failure or safety-critical probability for the family, $\mathrm{width}(\widehat{p}_{fail}(n))$ denotes the width of its uncertainty interval (for example, a confidence or credible interval), and $\epsilon_c,\epsilon_f,\epsilon_p$ are user-specified stopping tolerances. This stopping rule is operational: it states when additional WM rollouts have diminishing adequacy value for the tested planner and intent.

\subsection{Adequacy Metrics}

WM-Cov reports four families of metrics. First, requested-to-realized metrics measure whether the provider actually generated the conditions requested by the testing intent. Second, realized interaction coverage measures how much of the closed-loop behavior space was exercised. Third, evidence metrics count valid ADS failures, valid safety-critical events, and unique failure modes. Fourth, audit metrics count conditioning failures, invalid scenarios, duplicate failures, and artifact failures. The combination is necessary because high failure yield without coverage may overfit to one narrow failure type, while high coverage without validity may reward implausible or simulator-induced events.

\subsection{Adequacy Report}

The output of WM-Cov is not a single scalar score. For each scenario family, provider, ego planner, and budget, it produces an adequacy report
\begin{equation}
    \mathcal{A}_B =
    \{C_{req\rightarrow real},\, C_{valid},\, \mathrm{VF},\, \mathrm{FM},\, \mathrm{AI},\, \bar R,\, \bar Q,\, \Pi_v\},
\end{equation}
where $\mathrm{VF}$ is the number of valid ADS failures, $\mathrm{FM}$ is the number of unique failure modes, $\mathrm{AI}$ is the number of artifact, invalid, or conditioning-failure selections, $\bar R$ and $\bar Q$ are mean risk and realism, and $\Pi_v=\mathrm{VF}/(\mathrm{VF}+\mathrm{AI})$ is valid-evidence precision. A testing campaign is therefore reported as a convergence and coverage--validity--budget trade-off rather than as a raw count of generated failures. This adequacy report is the object interpreted by WM-Cov: additional rollouts are valuable only when they improve requested-to-valid coverage, valid evidence, failure-mode diversity, artifact burden, or uncertainty in the estimated valid-failure probability under the available budget.

\section{Method: WM-Cov}
\label{sec:method}

\begin{figure*}[t]
    \centering
    \resizebox{\textwidth}{!}{%
    \begin{tikzpicture}[
        font=\sffamily\footnotesize,
        card/.style={draw, rounded corners=2pt, align=center, minimum height=9mm, line width=0.55pt},
        bundle/.style={card, minimum width=23mm, text width=21mm},
        evidence/.style={card, minimum width=20mm, text width=18mm, font=\sffamily\scriptsize},
        arrowtext/.style={font=\Large, text=black!45},
        guide/.style={line width=0.38pt, draw=black!30},
        >=stealth
    ]

    \node[font=\bfseries\scriptsize, anchor=west] at (-8.18,1.58) {Input};
    \node[font=\bfseries\scriptsize, anchor=west] at (-5.44,1.58) {Interactive provider};
    \node[font=\bfseries\scriptsize, anchor=west] at (-0.92,1.58) {WM-Cov adequacy layer};
    \draw[guide] (-8.18,1.40) -- (-7.15,1.40);
    \draw[guide] (-5.44,1.40) -- (-2.52,1.40);
    \draw[guide] (-0.92,1.40) -- (8.40,1.40);

    \node[bundle, fill=blue!5, draw=blue!45!black] (req) at (-7.45,0.35)
    {\textbf{Test request}\\intent, ODD,\\ego planner};

    \node[bundle, fill=orange!8, draw=orange!70!black] (wm) at (-4.65,0.35)
    {\textbf{WM-style}\\closed-loop\\provider};

    \node[bundle, fill=orange!4, draw=orange!55!black] (trace) at (-1.85,0.35)
    {\textbf{Planner-conditioned}\\rollout traces};

    \node[bundle, fill=white, draw=black!42] (adapter) at (0.95,0.35)
    {\textbf{Trace adapter}\\request--realized\\bins};

    \node[bundle, fill=white, draw=black!42] (audit) at (3.75,0.35)
    {\textbf{Evidence audit}\\validity, duplicates,\\artifacts};

    \node[card, minimum width=25mm, text width=23mm, fill=yellow!8, draw=orange!65!black] (report) at (6.55,0.35)
    {\textbf{Stopping report}\\coverage growth,\\failures, precision};

    \node[arrowtext] at (-6.05,0.35) {$\rightarrow$};
    \node[arrowtext] at (-3.25,0.35) {$\rightarrow$};
    \node[arrowtext] at (-0.45,0.35) {$\rightarrow$};
    \node[arrowtext] at (2.35,0.35) {$\rightarrow$};
    \node[arrowtext] at (5.15,0.35) {$\rightarrow$};

    \node[evidence, fill=green!7, draw=green!45!black] (valid) at (1.05,-1.70) {valid evidence};
    \node[evidence, fill=gray!8, draw=black!35] (dup) at (3.75,-1.70) {duplicate accounting};
    \node[evidence, fill=red!6, draw=red!55!black] (art) at (6.45,-1.70) {artifact / invalid\\suppression};
    \node[font=\scriptsize, text=black!55] at (3.75,-2.35)
    {audited evidence categories before budgeted accumulation};

    \draw[-latex, dashed, line width=0.55pt, draw=orange!70!black]
    (report.south) .. controls (5.70,-1.05) and (-6.85,-1.05) ..
    node[pos=0.50, below, font=\scriptsize, text=orange!70!black]
    {campaign iteration: stop, continue, or reallocate budget}
    (req.south);

    \end{tikzpicture}
    }
    \caption{WM-Cov campaign loop for interactive testing. A testing intent, prompt/ODD condition, and ego planner are executed through a world-model-style closed-loop provider. WM-Cov converts provider artifacts into a trace-adapter schema, separates requested and realized bins, audits validity, suppresses duplicate and artifact evidence, and reports stopping-oriented adequacy under a finite budget. The report feeds the next campaign decision: stop when valid evidence has saturated, or continue by reallocating budget to under-realized prompts, seeds, routes, or horizons.}
    \label{fig:wm_cov_pipeline}
\end{figure*}
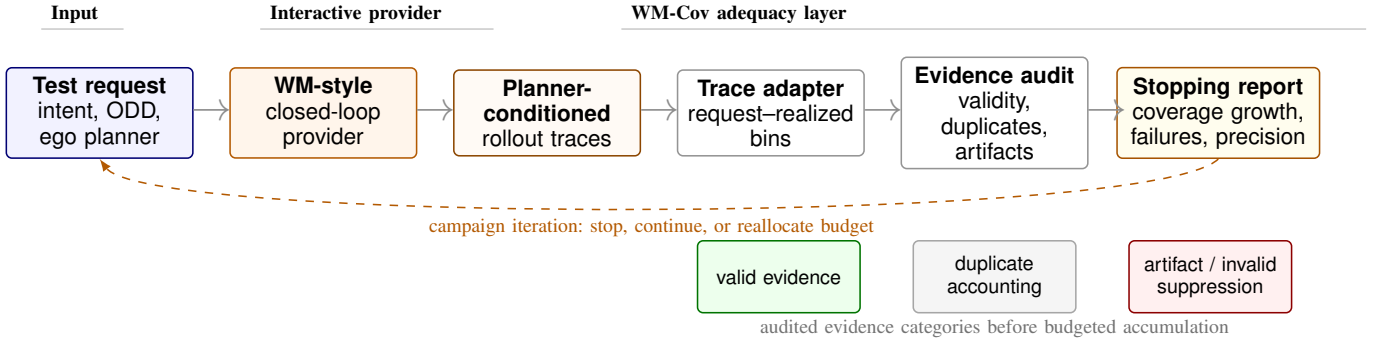

WM-Cov is an evaluation layer placed above a world model or generative simulator. It does not require modifying the generator. Instead, it standardizes closed-loop traces, audits whether requested conditions were realized, audits validity, and reports stopping-oriented adequacy.
Figure~\ref{fig:wm_cov_pipeline} gives the corresponding campaign view: testing requests and planners enter an interactive provider, while WM-Cov performs trace adaptation, evidence audit, duplicate/artifact accounting, and budgeted adequacy reporting. The report then feeds the next testing decision by stopping the campaign or allocating additional budget to prompts, seeds, routes, or horizons that still lack valid evidence.

\subsection{Trace Ingestion}

WM-Cov operates after an interactive testing provider has produced a finite set of planner-conditioned traces. This post-provider trace set is the common evidence object for adequacy accounting: it may come directly from a neural world model, a closed-loop generative simulator, a trajectory generator, rule perturbation, or log replay used as a controlled proxy for generated rollout evidence. The framework therefore accepts closed-loop traces through a common adapter contract. Each adapter exports scenario-family metadata, requested conditions, realized interaction metadata, actor trajectories when available, provider type, replay information, and execution notes. Downstream WM-Cov modules then compute requested-to-realized coverage, risk bins, failure modes, validity labels, and stopping metrics. This separation avoids embedding coverage logic inside a simulator-specific exporter.

The adapter contract is deliberately minimal. A provider is not required to expose its internal model states; it only needs to export the information required for replay, feature mapping, and evidence audit. Table~\ref{tab:adapter_schema} summarizes the required fields. Optional raw trajectories or sensor references can be retained for evidence audit, but the coverage and stopping modules operate on the standardized fields.

\begin{table}[t]
\centering
\caption{Provider-agnostic interactive trace adapter schema.}
\label{tab:adapter_schema}
\footnotesize
\begin{tabular}{@{}>{\raggedright\arraybackslash}p{0.26\columnwidth}>{\raggedright\arraybackslash}p{0.21\columnwidth}>{\raggedright\arraybackslash}p{0.39\columnwidth}@{}}
\toprule
Field & Type & Purpose \\
\midrule
trace\_id / source\_id & string & Stable identifier for replay and audit. \\
family\_id & string & Testing intent and scenario-family identifier. \\
ego\_planner\_id & string & Planner evaluated in the closed-loop trace. \\
provider & categorical & Log, simulator, perturbation engine, or world model. \\
seed\_id & string & Source log, route, map, prompt, or latent seed reference. \\
requested\_condition & structured & ODD bins requested by the testing intent. \\
realized\_interaction & structured & Ego/background behavior actually observed. \\
actors & structured & Ego, vehicles, pedestrians, and interaction metadata. \\
outcome & structured & Collision, near miss, hard braking, safe, or unknown. \\
scores & numeric & Risk, realism, artifact risk, and replay confidence. \\
audit\_note & text & Human- or rule-generated validity rationale. \\
\bottomrule
\end{tabular}
\end{table}

Different providers fill different parts of this schema. Video world models may return generated scenes and trajectory proxies; closed-loop simulators return executed episodes; trajectory generators return actor tracks; and log replay returns recorded seed coverage. WM-Cov requires replayable traces, requested and realized feature metadata, scores, and audit notes, not access to internal latent states.

\begin{figure*}[t]
\centering
\resizebox{\textwidth}{!}{%
\begin{tikzpicture}[
    font=\sffamily\footnotesize,
    block/.style={draw, rounded corners=2pt, align=center, minimum height=10mm, line width=0.55pt},
    main/.style={block, minimum width=34mm, text width=32mm},
    bin/.style={block, minimum width=35mm, text width=33mm, font=\sffamily\scriptsize},
    note/.style={draw, rounded corners=2pt, align=center, font=\scriptsize, minimum height=7mm, line width=0.48pt},
    arrow/.style={-latex, line width=0.58pt, draw=black!72, shorten >=1pt, shorten <=1pt},
    >=stealth
]

\node[font=\bfseries\small] at (0,2.72)
{Evidence accounting: raw danger $\rightarrow$ valid test evidence};

\node[main, fill=blue!5, draw=blue!45!black] (raw) at (-6.75,1.35)
{\textbf{Generated / simulated}\\candidates\\collisions, near misses,\\hard braking, safe cases};

\node[main, fill=gray!5, draw=black!45] (ledger) at (-2.35,1.35)
{\textbf{Feature mapping}\\and validity ledger\\ODD, maneuver, actors,\\outcome, realism, audit tag};

\node[main, fill=green!6, draw=green!45!black] (selector) at (2.35,1.35)
{\textbf{Budgeted selector}\\coverage, risk, realism\\minus duplicate\\and artifact cost};

\node[main, fill=yellow!8, draw=orange!65!black] (report) at (6.75,1.35)
{\textbf{Adequacy report}\\coverage, VF, FM, AI\\valid-evidence precision\\$\Pi_v=VF/(VF+AI)$};

\draw[arrow] (raw.east) -- (ledger.west);
\draw[arrow] (ledger.east) -- (selector.west);
\draw[arrow] (selector.east) -- (report.west);

\node[font=\scriptsize, text=black!55] at (0,0.22)
{validity categories recorded before budgeted selection};

\node[bin, fill=green!7, draw=green!45!black] (valid) at (-4.70,-0.48)
{\textbf{Valid evidence}\\counts toward coverage,\\valid failures, modes};

\node[bin, fill=gray!8, draw=black!38] (dup) at (0,-0.48)
{\textbf{Duplicate evidence}\\tracked in ledger,\\not new diversity};

\node[bin, fill=red!7, draw=red!55!black] (artifact) at (4.70,-0.48)
{\textbf{Artifact / invalid}\\penalized and excluded\\from valid-failure yield};

\node[note, fill=red!4, draw=red!45!black, text width=43mm] (danger) at (-4.70,-1.80)
{Dangerous-looking outputs are not automatically counted as testing evidence.};

\node[note, fill=green!4, draw=green!45!black, text width=43mm] (filter) at (4.70,-1.80)
{Filters increase useful evidence density under a fixed test budget.};

\draw[arrow] (danger.north) -- (valid.south);
\draw[arrow] (filter.north) -- (artifact.south);

\node[note, fill=blue!3, draw=blue!35!black, text width=43mm] at (-4.70,-3.02)
{\textbf{Executed pool:} 1219 events contain 529 valid events and 690 artifact failures.};

\node[note, fill=blue!3, draw=blue!35!black, text width=43mm] at (0,-3.02)
{\textbf{WM-like pool:} 1500 candidates contain valid failures, critical cases, duplicates, and artifacts.};

\node[note, fill=green!4, draw=green!45!black, text width=43mm] at (4.70,-3.02)
{\textbf{WM-Cov at $B=100$:} 99/100 and 76/100 selected candidates are valid failures, with zero artifacts.};

\end{tikzpicture}
}
\caption{Evidence accounting view of WM-Cov. A world model or generative simulator may produce many dangerous-looking candidates, but WM-Cov treats them as test evidence only after feature mapping, validity auditing, duplicate/artifact accounting, and budgeted adequacy selection. The resulting report summarizes valid, duplicate, and artifact/invalid evidence so that coverage and valid-evidence precision are reported under a fixed testing budget rather than from raw dangerous-looking outputs.}
\label{fig:evidence_flow}
\end{figure*}
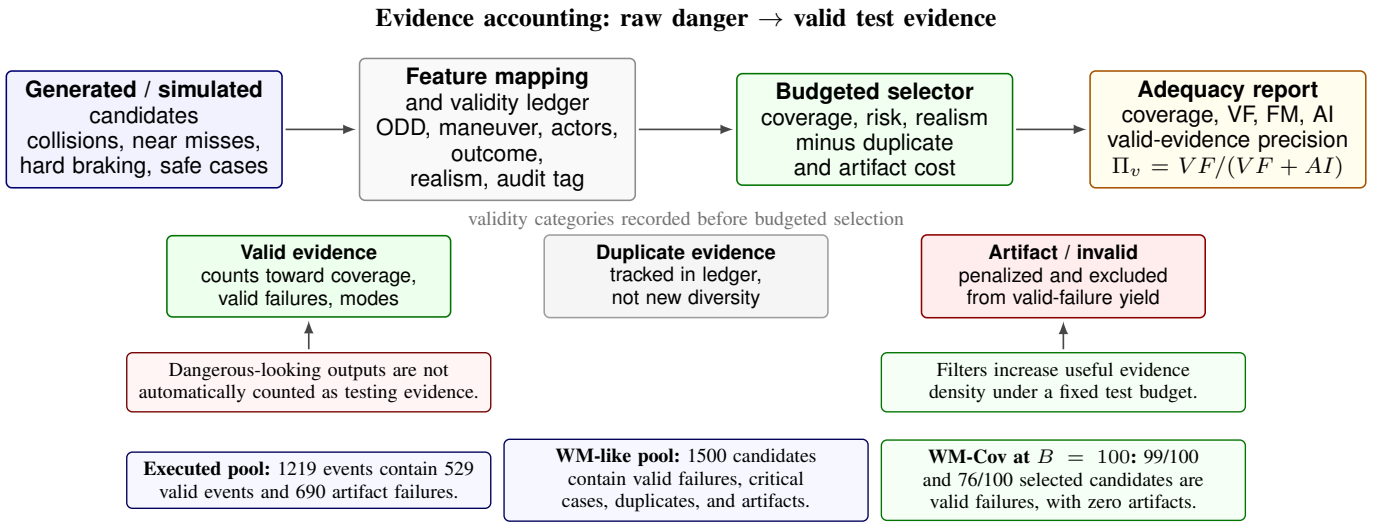

Figure~\ref{fig:evidence_flow} summarizes the evidence-accounting view used by WM-Cov. A generated or simulated trace is not counted as evidence merely because it contains a collision, near miss, hard braking event, or high-risk score. It first enters a validity ledger, and only audited traces are used for budgeted accumulation and adequacy reporting.

\subsection{Feature Taxonomy}

The taxonomy includes ODD, interaction, risk bin, behavior, failure mode, and validity. These dimensions are selected because they jointly describe where the scenario occurs, how agents interact, why it is risky, what behavior is expressed, what outcome occurs, and whether the outcome is admissible as test evidence.

For a selected set $\mathcal{T}$, single-dimension coverage is computed over each categorical dimension $d$. Let $\phi_d(\tau)$ denote the projection of the trace feature vector $\phi(\tau)$ onto dimension $d$, and let $\Omega_d$ be the set of admissible bins for that dimension:
\begin{equation}
    \mathrm{Cov}_{d}(\mathcal{T}) =
    \frac{|\{ \phi_d(\tau): \tau\in\mathcal{T}\}\cap\Omega_d|}{|\Omega_d|},
\end{equation}
Pairwise coverage $\mathrm{Cov}_{pair}(\mathcal{T})$ is computed analogously over selected dimension pairs, such as interaction--risk and behavior--failure. The final adequacy coverage combines weighted single-dimension and pairwise coverage:
\begin{equation}
    \mathrm{Cov}(\mathcal{T}) =
    \alpha \sum_d w_d \mathrm{Cov}_{d}(\mathcal{T})
    + (1-\alpha)\mathrm{Cov}_{pair}(\mathcal{T}).
\end{equation}
Here $w_d$ is the weight assigned to dimension $d$, normalized so that $\sum_d w_d=1$, and $\alpha\in[0,1]$ controls the balance between single-dimension and pairwise coverage.
The current implementation uses both single-bin and pairwise-bin coverage because autonomous-driving failures often arise from interactions between context and behavior rather than from isolated categories.

\subsection{Validity Audit}

WM-Cov separates valid ADS failures, valid safety-critical events, duplicate failures, conditioning failures, invalid scenarios, and artifact failures. This distinction is central: a dangerous-looking generated event should not automatically count as test evidence. For example, a collision caused by plausible non-yielding behavior should be treated differently from an emergency stop caused by a disconnected map edge. Similarly, a trace requested as a rainy night pedestrian crossing should not count toward that bin if the realized trace is a daytime vehicle-only interaction. The audit layer records both the label and a short audit reason so that results can be inspected rather than only aggregated. Table~\ref{tab:validity_ledger} defines the ledger categories used before traces are admitted to budgeted adequacy selection.

\begin{table}[t]
\centering
\caption{Validity ledger used by WM-Cov before budgeted selection.}
\label{tab:validity_ledger}
\footnotesize
\begin{tabular}{@{}>{\raggedright\arraybackslash}p{0.24\columnwidth}>{\raggedright\arraybackslash}p{0.14\columnwidth}>{\raggedright\arraybackslash}p{0.42\columnwidth}@{}}
\toprule
Label & Evidence & Operational rule \\
\midrule
Valid ADS failure & yes & Plausible, replayable, ADS-relevant failure outcome. \\
Valid safety-critical & yes & Plausible high-risk event without confirmed ADS failure. \\
Valid safe & yes & Plausible non-critical case useful for coverage accounting. \\
Duplicate failure & limited & Repeated failure pattern counted once for diversity-sensitive metrics. \\
Conditioning failure & no & Requested ODD or interaction condition was not realized. \\
Artifact / invalid & no & Outcome caused by generation, simulator, map, or dynamics artifact. \\
\bottomrule
\end{tabular}
\end{table}

\subsection{Budgeted Selection}

The multi-objective selection policy is implemented greedily when a provider offers a pool of generated or replayable traces. For an online world-model provider, the same budgeted view becomes adaptive rollout allocation: the selected unit can be the next scenario family, prompt, seed, route, horizon, or planner-conditioned request rather than only an already generated trace. At each step, WM-Cov selects the trace or rollout request with the largest score:
\begin{align}
    s(\tau|\mathcal{T}) ={}&
    \beta_r R(\tau) + \beta_q Q(\tau)
    + \beta_c \Delta C_{valid}(\tau|\mathcal{T}) \nonumber\\
    &- \beta_m M(\tau|\mathcal{T}) - \beta_a A(\tau),
\end{align}
where $\Delta C_{valid}$ is the marginal gain in valid realized coverage, $M$ is redundancy with already selected traces, $A$ is artifact risk, and $\beta_r,\beta_q,\beta_c,\beta_m,\beta_a$ are nonnegative selection weights. Baselines include random, risk-only, realism-only, and coverage-only selection. Comparing these baselines exposes whether a testing strategy is merely finding failures, merely spreading across the requested feature space, or actually producing valid and diverse interaction evidence.

The selector is an operational policy; the main contribution is the evidence-accounting and stopping protocol around the adapter, requested/realized coverage ledger, validity ledger, duplicate/artifact accounting, and adequacy report. Table~\ref{tab:wmcov_algorithm} summarizes this procedure from test-intent specification through stopping-oriented reporting.

\begin{table}[t]
\centering
\caption{WM-Cov interactive adequacy assessment procedure.}
\label{tab:wmcov_algorithm}
\footnotesize
\begin{tabular}{@{}r p{0.82\columnwidth}@{}}
\toprule
Step & Operation \\
\midrule
1 & Specify testing intent $q$, scenario family $\mathcal{F}(q)$, ego planner $\pi_e$, and requested coverage bins $\Omega_{req}(q)$. \\
2 & Execute or ingest closed-loop traces through the adapter schema in Table~\ref{tab:adapter_schema}. \\
3 & Map each trace to $\phi(\tau_i)$ and compute $R(\tau_i)$, $Q(\tau_i)$, and $A(\tau_i)$. \\
4 & Assign conditioning and validity labels using the ledger in Table~\ref{tab:validity_ledger}. \\
5 & Accumulate traces under budget using $s(\tau|\mathcal{T})$ or a provider sampling policy. \\
6 & Update requested-to-realized coverage, valid realized coverage, duplicate counts, artifact counts, and failure-mode discovery curves. \\
7 & Stop when coverage growth, valid failure discovery, and uncertainty satisfy the stopping criteria, or report $\mathcal{A}_B$ at the budget limit. \\
\bottomrule
\end{tabular}
\end{table}

\section{Evaluation Studies}
\label{sec:experiments}

The evaluation studies test the evidence-accounting components required by the interactive adequacy framework. They are scoped as an adequacy-layer validation rather than as a comprehensive multi-map, multi-prompt, multi-planner neural world-model benchmark. Specifically, they ask whether WM-Cov separates dangerous-looking traces from valid evidence, improves coverage--risk--realism--artifact trade-offs under budget, remains stable across selector weights, and converts real DriveArena TrafficManager--WorldDreamer outputs into requested-versus-realized evidence records. The final DriveArena study covers two ego planners, two horizons, and six prompt conditions, with 360 total ego-route requests, followed by an 80-request disjoint route-slice check. Together, these studies evaluate whether the proposed ledger, budgeted selection rule, and stopping report provide usable adequacy evidence for interactive world-model-style testing.

\subsection{Trace Sources}

We first use two finite trace-evidence sources to test the accounting layer before applying the same schema to DriveArena provider artifacts. The first source consists of executed TeraSim/SUMO event-level traces extracted from simulator logs. It is useful because each trace has an executed outcome, including collisions, hard braking, and simulator-level warnings.

The second source is a WM-like mixed trace pool constructed to emulate the evidence profile expected from world-model-based testing. It contains plausible safety-critical cases, valid ADS failures, duplicate failures, and artifact-like failures, allowing us to stress-test WM-Cov when a provider returns both useful and misleading dangerous-looking traces.

For both pools, traces are mapped to the same feature schema: ODD, interaction type, risk bin, behavior class, failure mode, and validity label. In the full interactive setting, these fields correspond to realized trace attributes and are compared against requested bins from the testing intent. This shared representation lets us compare selection strategies without binding the evaluation logic to a particular generator.

Thus, the experiments evaluate the WM-Cov adequacy layer under executed generative-simulator, WM-like, and scoped provider sources. The DriveArena provider matrix verifies that generated sensor frames, ego-planner responses, partial realizations, and condition-specific failures can be ingested as planner-conditioned realized interaction traces and reported as requested-to-realized evidence.

\subsection{DriveArena Adapter and Closed-Loop Provider Batch}

DriveArena is the real provider used for the scoped interactive scenario-family study because its architecture separates a traffic manager, a WorldDreamer sensor generator, and an ego driving agent. In this setup, TrafficManager specifies maps, routes, background traffic, collision and off-road checks, and structured run logs; WorldDreamer generates sensor observations; and a driving agent such as UniAD or VAD produces the ego trajectory. Replacing the driving agent can therefore change the realized interaction trace, which is precisely the setting targeted by WM-Cov.

The DriveArena adapter maps \texttt{drive\_arena.pkl} artifacts to the WM-Cov schema and records the provider, scenario family, route, ego id, realized actors, route completion, checker events, replay artifact, risk proxy, route edges, and run status.

Adapter feasibility is checked on TrafficManager artifacts and a SUMO/TraCI fallback over DriveArena's boston-seaport network. Native \texttt{drive\_arena.pkl} files and route-level traces are converted into WM-Cov rows, while the main provider-chain evidence comes from the balanced TrafficManager--WorldDreamer matrix below.

\begin{figure*}[t]
    \centering
    \includegraphics[width=\textwidth]{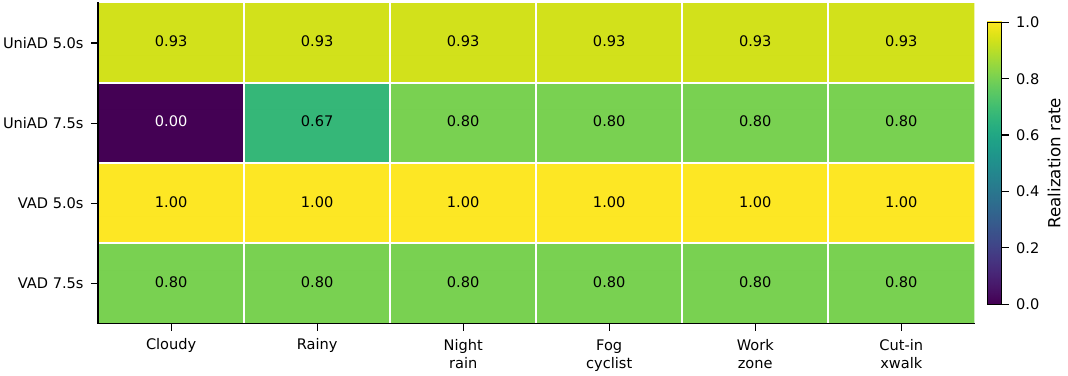}
    \caption{Requested-to-realized accounting in the DriveArena TrafficManager--WorldDreamer six-prompt matrix. Each heatmap cell aggregates 15 requested closed-loop ego-route attempts for one planner, prompt, and horizon. Fully realized traces and pkl-backed partial realizations are distinct evidence objects in WM-Cov. The matrix shows that realization is planner-, horizon-, and prompt-conditioned; raw requested counts would hide the UniAD default-cloudy 7.5-s provider-chain failure and the repeated 7.5-s partial-realization effect.}
    \label{fig:drivearena_formal_matrix}
\end{figure*}

We ran a balanced provider matrix with TrafficManager, WorldDreamer, and two ego planners, UniAD and VAD. The matrix uses two horizons, 5.0 s and 7.5 s, and six prompt-description conditions: \emph{default cloudy}, \emph{rainy dusk}, \emph{night rain occlusion}, \emph{fog dawn cyclists}, \emph{construction lane closure}, and \emph{aggressive cut-in with crosswalk}. Each planner--horizon--prompt cell contains 15 requested ego-route attempts, yielding 360 requested closed-loop attempts in total. Of these, 304 produce fully realized \texttt{drive\_arena.pkl}-backed closed-loop traces, 56 produce pkl-backed partial realizations, and none are completely unrealized. The matrix exposes the accounting problem that WM-Cov is designed to measure. VAD realizes all 5.0-s traces and 72 of 90 requested 7.5-s traces, whereas UniAD realizes 84 of 90 requested 5.0-s traces and 58 of 90 requested 7.5-s traces. The UniAD default-cloudy 7.5-s cell is the clearest provider-chain realization failure: it produces partial artifacts but no UniAD planner responses, so WM-Cov counts it as partial provider evidence rather than as planner safety evidence. Across the remaining prompt families, the matrix shows a repeated horizon effect: 5.0-s cells are mostly realized, while 7.5-s cells more often expose partial provider-chain realization. Thus, the result is not a safety ranking between planners; it is evidence that realization itself is planner-, horizon-, and prompt-conditioned. Because reference-image conditioning is not independently evaluated, these results are used as provider-chain and request-realization evidence, not as a visual-realism benchmark, ADS failure result, or adequacy-sufficiency claim. The provider-chain evidence is summarized in Table~\ref{tab:drivearena_closed_loop_batch}, Fig.~\ref{fig:drivearena_formal_matrix}, and Table~\ref{tab:drivearena_second_route_slice}.

\begin{table}[t]
\centering
\caption{DriveArena TrafficManager--WorldDreamer six-prompt provider-accounting summary.}
\label{tab:drivearena_closed_loop_batch}
\scriptsize
\setlength{\tabcolsep}{1.8pt}
\begin{tabular}{@{}llrrrr@{}}
\toprule
Planner & Setting & Req. & Real. & Part. & Rate \\
\midrule
UniAD & 5.0 s, six prompts & 90 & 84 & 6 & 0.933 \\
UniAD & 7.5 s, six prompts & 90 & 58 & 32 & 0.644 \\
VAD & 5.0 s, six prompts & 90 & 90 & 0 & 1.000 \\
VAD & 7.5 s, six prompts & 90 & 72 & 18 & 0.800 \\
\midrule
Total & all cells & 360 & 304 & 56 & 0.844 \\
\bottomrule
\end{tabular}
\end{table}

\IfFileExists{figures/drivearena_second_route_slice/second_route_slice_check.tex}{%
\begin{table}[t]
\centering
\caption{Disjoint route-slice check on DriveArena TrafficManager--WorldDreamer.}
\label{tab:drivearena_second_route_slice}
\begin{tabular}{llrrrr}
\toprule
Planner & Horizon & Req. & Full & Partial & Rate \\
\midrule
UniAD & 5.0 s & 20 & 18 & 2 & 0.900 \\
UniAD & 7.5 s & 20 & 18 & 2 & 0.900 \\
VAD & 5.0 s & 20 & 20 & 0 & 1.000 \\
VAD & 7.5 s & 20 & 18 & 2 & 0.900 \\
\midrule
Total & all & 80 & 74 & 6 & 0.925 \\
\bottomrule
\end{tabular}
\end{table}

Table~\ref{tab:drivearena_second_route_slice} reports a route-slice robustness check. We re-instantiated the same two prompt conditions (default cloudy, rainy dusk) on a disjoint ego-route slice, ego ids 16--25, with the same two planners and two horizons. Across 80 additional requested closed-loop attempts, 74 became fully realized traces, 6 remained pkl-backed partial realizations, and 0 were completely unrealized. The condition-aware UniAD--VAD ledger contains 40 matched route rows, of which 36 are jointly fully realized. This check strengthens the interpretation that requested-to-realized accounting captures provider-chain behavior rather than a route-specific artifact, with broader multi-map evaluation remaining an important next step.
}{}

\subsection{Baselines}

We compare WM-Cov with five budgeted baselines. Random selection estimates unguided sampling. Scenario-tag selection greedily covers curated tags without risk, realism, or artifact penalties. Risk-only selection approximates failure hunting. Realism-only selection prioritizes validity-confidence scores. Coverage-only selection prioritizes marginal feature coverage without artifact penalties. WM-Cov jointly scores marginal coverage, risk, realism, redundancy, and artifact risk.

All methods are evaluated at budget 100. For the executed TeraSim/SUMO pool, we also report a budget curve to show how coverage and evidence accumulate as the budget increases.

\subsection{Metrics}

Metrics include coverage, single-feature coverage, pairwise feature coverage, valid ADS failures, valid safety-critical events, unique failure modes, invalid/artifact failures, mean risk, and mean realism. We interpret artifact/invalid counts as negative evidence: they represent generated or simulated events that look critical but should not be counted as valid ADS test failures.

\subsection{Sensitivity Protocol}

To test whether WM-Cov depends on a narrow weight setting, we run a grid over coverage and artifact-penalty weights. The sensitivity analysis reports the range of selected coverage, valid failures, and artifact/invalid counts across the grid. Stable ranges indicate that the core selection behavior is driven by the trace-evidence structure rather than by one fragile coefficient choice.

We also run an artifact-contamination stress test on the WM-like pool. For each contamination level, we construct fixed-size trace pools by mixing valid traces with artifact-like or duplicate traces and repeat the selection three times. This test approximates a world-model provider whose generated pool contains an increasing fraction of misleading dangerous-looking cases.

\subsection{Author-Verified Validity Audit}

Because validity labels are central to WM-Cov, we conduct an author-verified audit on a stratified 141-row blind sample covering valid failures, safety-critical events, duplicates, artifacts, and safe cases. Initial labels are assigned using a written rubric and then reviewed by the authors for all disagreement cases. We report exact agreement, Cohen's kappa, and artifact precision/recall. The stricter duplicate rule is diagnostic only: full-pool duplicate evidence requires source-specific keys such as event actor, lane, time, or generated-scenario identifier.

\FloatBarrier
\section{Results}
\label{sec:results}

Table~\ref{tab:event_distribution} summarizes the executed TeraSim/SUMO event pool. The pool contains 1219 event-level traces, including 230 valid ADS failures, 299 valid safety-critical events, and 690 artifact failures. The large artifact count is important: if every collision warning or emergency stop were treated as test evidence, the testing campaign would substantially overstate its safety relevance.

Table~\ref{tab:budget_100} shows the budget-100 selection trade-off on this executed pool. Risk-only and realism-only selection both select 100 valid failures with zero artifacts, but their coverage is narrow. Scenario-tag and coverage-only selection reach the highest coverage among these policies, but they admit 58 artifact or invalid cases and find only 17 valid failures. WM-Cov selects 99 valid failures with zero artifacts while retaining substantially more coverage than risk-only and realism-only selection. In this executed-pool evaluation, adequacy assessment therefore depends on the joint coverage--validity trade-off rather than on failure count or coverage alone.

\begin{figure*}[t]
\centering
\begin{tikzpicture}[font=\sffamily]
\begin{scope}
\draw[->, line width=0.35pt] (0,0.35) -- (4.25,0.35);
\draw[->, line width=0.35pt] (0,0.35) -- (0,3.55);
\node[font=\bfseries\footnotesize, anchor=west] at (0,3.78) {(a) Coverage};
\draw (0.400,0.30) -- (0.400,0.40);
\node[font=\scriptsize, anchor=north] at (0.400,0.25) {20};
\draw (1.200,0.30) -- (1.200,0.40);
\node[font=\scriptsize, anchor=north] at (1.200,0.25) {60};
\draw (2.000,0.30) -- (2.000,0.40);
\node[font=\scriptsize, anchor=north] at (2.000,0.25) {100};
\draw (3.000,0.30) -- (3.000,0.40);
\node[font=\scriptsize, anchor=north] at (3.000,0.25) {150};
\draw (4.000,0.30) -- (4.000,0.40);
\node[font=\scriptsize, anchor=north] at (4.000,0.25) {200};
\draw (-0.05,0.683) -- (0.05,0.683);
\node[font=\scriptsize, anchor=east] at (-0.08,0.683) {0.10};
\draw[gray!16, line width=0.25pt] (0,0.683) -- (4.05,0.683);
\draw (-0.05,1.517) -- (0.05,1.517);
\node[font=\scriptsize, anchor=east] at (-0.08,1.517) {0.15};
\draw[gray!16, line width=0.25pt] (0,1.517) -- (4.05,1.517);
\draw (-0.05,2.350) -- (0.05,2.350);
\node[font=\scriptsize, anchor=east] at (-0.08,2.350) {0.20};
\draw[gray!16, line width=0.25pt] (0,2.350) -- (4.05,2.350);
\draw (-0.05,3.183) -- (0.05,3.183);
\node[font=\scriptsize, anchor=east] at (-0.08,3.183) {0.25};
\draw[gray!16, line width=0.25pt] (0,3.183) -- (4.05,3.183);
\draw[blue!70!black, solid, line width=0.7pt] (0.400,2.832) -- (0.800,2.832) -- (1.200,2.832) -- (2.000,2.832) -- (3.000,2.832) -- (4.000,2.832);
\node[blue!70!black, scale=0.45] at (0.400,2.832) {$\bullet$};
\node[blue!70!black, scale=0.45] at (0.800,2.832) {$\bullet$};
\node[blue!70!black, scale=0.45] at (1.200,2.832) {$\bullet$};
\node[blue!70!black, scale=0.45] at (2.000,2.832) {$\bullet$};
\node[blue!70!black, scale=0.45] at (3.000,2.832) {$\bullet$};
\node[blue!70!black, scale=0.45] at (4.000,2.832) {$\bullet$};
\draw[red!70!black, dashed, line width=0.7pt] (0.400,0.557) -- (0.800,0.557) -- (1.200,0.557) -- (2.000,0.557) -- (3.000,1.515) -- (4.000,1.515);
\node[red!70!black, scale=0.45] at (0.400,0.557) {$\circ$};
\node[red!70!black, scale=0.45] at (0.800,0.557) {$\circ$};
\node[red!70!black, scale=0.45] at (1.200,0.557) {$\circ$};
\node[red!70!black, scale=0.45] at (2.000,0.557) {$\circ$};
\node[red!70!black, scale=0.45] at (3.000,1.515) {$\circ$};
\node[red!70!black, scale=0.45] at (4.000,1.515) {$\circ$};
\draw[green!45!black, dash dot, line width=0.7pt] (0.400,3.060) -- (0.800,3.060) -- (1.200,3.060) -- (2.000,3.060) -- (3.000,3.060) -- (4.000,3.060);
\node[green!45!black, scale=0.45] at (0.400,3.060) {$\triangle$};
\node[green!45!black, scale=0.45] at (0.800,3.060) {$\triangle$};
\node[green!45!black, scale=0.45] at (1.200,3.060) {$\triangle$};
\node[green!45!black, scale=0.45] at (2.000,3.060) {$\triangle$};
\node[green!45!black, scale=0.45] at (3.000,3.060) {$\triangle$};
\node[green!45!black, scale=0.45] at (4.000,3.060) {$\triangle$};
\draw[black!65, dotted, line width=0.7pt] (0.400,2.957) -- (0.800,2.957) -- (1.200,2.957) -- (2.000,2.957) -- (3.000,3.060) -- (4.000,3.060);
\node[black!65, scale=0.45] at (0.400,2.957) {$\diamond$};
\node[black!65, scale=0.45] at (0.800,2.957) {$\diamond$};
\node[black!65, scale=0.45] at (1.200,2.957) {$\diamond$};
\node[black!65, scale=0.45] at (2.000,2.957) {$\diamond$};
\node[black!65, scale=0.45] at (3.000,3.060) {$\diamond$};
\node[black!65, scale=0.45] at (4.000,3.060) {$\diamond$};
\end{scope}
\begin{scope}[xshift=5.35cm]
\draw[->, line width=0.35pt] (0,0.35) -- (4.25,0.35);
\draw[->, line width=0.35pt] (0,0.35) -- (0,3.55);
\node[font=\bfseries\footnotesize, anchor=west] at (0,3.78) {(b) Valid failures};
\draw (0.400,0.30) -- (0.400,0.40);
\node[font=\scriptsize, anchor=north] at (0.400,0.25) {20};
\draw (1.200,0.30) -- (1.200,0.40);
\node[font=\scriptsize, anchor=north] at (1.200,0.25) {60};
\draw (2.000,0.30) -- (2.000,0.40);
\node[font=\scriptsize, anchor=north] at (2.000,0.25) {100};
\draw (3.000,0.30) -- (3.000,0.40);
\node[font=\scriptsize, anchor=north] at (3.000,0.25) {150};
\draw (4.000,0.30) -- (4.000,0.40);
\node[font=\scriptsize, anchor=north] at (4.000,0.25) {200};
\draw (-0.05,0.350) -- (0.05,0.350);
\node[font=\scriptsize, anchor=east] at (-0.08,0.350) {0};
\draw[gray!16, line width=0.25pt] (0,0.350) -- (4.05,0.350);
\draw (-0.05,1.100) -- (0.05,1.100);
\node[font=\scriptsize, anchor=east] at (-0.08,1.100) {50};
\draw[gray!16, line width=0.25pt] (0,1.100) -- (4.05,1.100);
\draw (-0.05,1.850) -- (0.05,1.850);
\node[font=\scriptsize, anchor=east] at (-0.08,1.850) {100};
\draw[gray!16, line width=0.25pt] (0,1.850) -- (4.05,1.850);
\draw (-0.05,2.600) -- (0.05,2.600);
\node[font=\scriptsize, anchor=east] at (-0.08,2.600) {150};
\draw[gray!16, line width=0.25pt] (0,2.600) -- (4.05,2.600);
\draw (-0.05,3.350) -- (0.05,3.350);
\node[font=\scriptsize, anchor=east] at (-0.08,3.350) {200};
\draw[gray!16, line width=0.25pt] (0,3.350) -- (4.05,3.350);
\draw[blue!70!black, solid, line width=0.7pt] (0.400,0.635) -- (0.800,0.935) -- (1.200,1.235) -- (2.000,1.835) -- (3.000,2.585) -- (4.000,3.335);
\node[blue!70!black, scale=0.45] at (0.400,0.635) {$\bullet$};
\node[blue!70!black, scale=0.45] at (0.800,0.935) {$\bullet$};
\node[blue!70!black, scale=0.45] at (1.200,1.235) {$\bullet$};
\node[blue!70!black, scale=0.45] at (2.000,1.835) {$\bullet$};
\node[blue!70!black, scale=0.45] at (3.000,2.585) {$\bullet$};
\node[blue!70!black, scale=0.45] at (4.000,3.335) {$\bullet$};
\draw[red!70!black, dashed, line width=0.7pt] (0.400,0.650) -- (0.800,0.950) -- (1.200,1.250) -- (2.000,1.850) -- (3.000,2.600) -- (4.000,3.350);
\node[red!70!black, scale=0.45] at (0.400,0.650) {$\circ$};
\node[red!70!black, scale=0.45] at (0.800,0.950) {$\circ$};
\node[red!70!black, scale=0.45] at (1.200,1.250) {$\circ$};
\node[red!70!black, scale=0.45] at (2.000,1.850) {$\circ$};
\node[red!70!black, scale=0.45] at (3.000,2.600) {$\circ$};
\node[red!70!black, scale=0.45] at (4.000,3.350) {$\circ$};
\draw[green!45!black, dash dot, line width=0.7pt] (0.400,0.425) -- (0.800,0.455) -- (1.200,0.485) -- (2.000,0.605) -- (3.000,0.665) -- (4.000,0.890);
\node[green!45!black, scale=0.45] at (0.400,0.425) {$\triangle$};
\node[green!45!black, scale=0.45] at (0.800,0.455) {$\triangle$};
\node[green!45!black, scale=0.45] at (1.200,0.485) {$\triangle$};
\node[green!45!black, scale=0.45] at (2.000,0.605) {$\triangle$};
\node[green!45!black, scale=0.45] at (3.000,0.665) {$\triangle$};
\node[green!45!black, scale=0.45] at (4.000,0.890) {$\triangle$};
\draw[black!65, dotted, line width=0.7pt] (0.400,0.425) -- (0.800,0.470) -- (1.200,0.470) -- (2.000,0.575) -- (3.000,0.710) -- (4.000,0.875);
\node[black!65, scale=0.45] at (0.400,0.425) {$\diamond$};
\node[black!65, scale=0.45] at (0.800,0.470) {$\diamond$};
\node[black!65, scale=0.45] at (1.200,0.470) {$\diamond$};
\node[black!65, scale=0.45] at (2.000,0.575) {$\diamond$};
\node[black!65, scale=0.45] at (3.000,0.710) {$\diamond$};
\node[black!65, scale=0.45] at (4.000,0.875) {$\diamond$};
\end{scope}
\begin{scope}[xshift=10.7cm]
\draw[->, line width=0.35pt] (0,0.35) -- (4.25,0.35);
\draw[->, line width=0.35pt] (0,0.35) -- (0,3.55);
\node[font=\bfseries\footnotesize, anchor=west] at (0,3.78) {(c) Artifact/invalid};
\draw (0.400,0.30) -- (0.400,0.40);
\node[font=\scriptsize, anchor=north] at (0.400,0.25) {20};
\draw (1.200,0.30) -- (1.200,0.40);
\node[font=\scriptsize, anchor=north] at (1.200,0.25) {60};
\draw (2.000,0.30) -- (2.000,0.40);
\node[font=\scriptsize, anchor=north] at (2.000,0.25) {100};
\draw (3.000,0.30) -- (3.000,0.40);
\node[font=\scriptsize, anchor=north] at (3.000,0.25) {150};
\draw (4.000,0.30) -- (4.000,0.40);
\node[font=\scriptsize, anchor=north] at (4.000,0.25) {200};
\draw (-0.05,0.350) -- (0.05,0.350);
\node[font=\scriptsize, anchor=east] at (-0.08,0.350) {0};
\draw[gray!16, line width=0.25pt] (0,0.350) -- (4.05,0.350);
\draw (-0.05,1.100) -- (0.05,1.100);
\node[font=\scriptsize, anchor=east] at (-0.08,1.100) {30};
\draw[gray!16, line width=0.25pt] (0,1.100) -- (4.05,1.100);
\draw (-0.05,1.850) -- (0.05,1.850);
\node[font=\scriptsize, anchor=east] at (-0.08,1.850) {60};
\draw[gray!16, line width=0.25pt] (0,1.850) -- (4.05,1.850);
\draw (-0.05,2.600) -- (0.05,2.600);
\node[font=\scriptsize, anchor=east] at (-0.08,2.600) {90};
\draw[gray!16, line width=0.25pt] (0,2.600) -- (4.05,2.600);
\draw (-0.05,3.350) -- (0.05,3.350);
\node[font=\scriptsize, anchor=east] at (-0.08,3.350) {120};
\draw[gray!16, line width=0.25pt] (0,3.350) -- (4.05,3.350);
\draw[blue!70!black, solid, line width=0.7pt] (0.400,0.350) -- (0.800,0.350) -- (1.200,0.350) -- (2.000,0.350) -- (3.000,0.350) -- (4.000,0.350);
\node[blue!70!black, scale=0.45] at (0.400,0.350) {$\bullet$};
\node[blue!70!black, scale=0.45] at (0.800,0.350) {$\bullet$};
\node[blue!70!black, scale=0.45] at (1.200,0.350) {$\bullet$};
\node[blue!70!black, scale=0.45] at (2.000,0.350) {$\bullet$};
\node[blue!70!black, scale=0.45] at (3.000,0.350) {$\bullet$};
\node[blue!70!black, scale=0.45] at (4.000,0.350) {$\bullet$};
\draw[red!70!black, dashed, line width=0.7pt] (0.400,0.350) -- (0.800,0.350) -- (1.200,0.350) -- (2.000,0.350) -- (3.000,0.350) -- (4.000,0.350);
\node[red!70!black, scale=0.45] at (0.400,0.350) {$\circ$};
\node[red!70!black, scale=0.45] at (0.800,0.350) {$\circ$};
\node[red!70!black, scale=0.45] at (1.200,0.350) {$\circ$};
\node[red!70!black, scale=0.45] at (2.000,0.350) {$\circ$};
\node[red!70!black, scale=0.45] at (3.000,0.350) {$\circ$};
\node[red!70!black, scale=0.45] at (4.000,0.350) {$\circ$};
\draw[green!45!black, dash dot, line width=0.7pt] (0.400,0.675) -- (0.800,1.125) -- (1.200,1.350) -- (2.000,1.800) -- (3.000,2.575) -- (4.000,3.175);
\node[green!45!black, scale=0.45] at (0.400,0.675) {$\triangle$};
\node[green!45!black, scale=0.45] at (0.800,1.125) {$\triangle$};
\node[green!45!black, scale=0.45] at (1.200,1.350) {$\triangle$};
\node[green!45!black, scale=0.45] at (2.000,1.800) {$\triangle$};
\node[green!45!black, scale=0.45] at (3.000,2.575) {$\triangle$};
\node[green!45!black, scale=0.45] at (4.000,3.175) {$\triangle$};
\draw[black!65, dotted, line width=0.7pt] (0.400,0.675) -- (0.800,0.875) -- (1.200,1.175) -- (2.000,1.725) -- (3.000,2.350) -- (4.000,3.025);
\node[black!65, scale=0.45] at (0.400,0.675) {$\diamond$};
\node[black!65, scale=0.45] at (0.800,0.875) {$\diamond$};
\node[black!65, scale=0.45] at (1.200,1.175) {$\diamond$};
\node[black!65, scale=0.45] at (2.000,1.725) {$\diamond$};
\node[black!65, scale=0.45] at (3.000,2.350) {$\diamond$};
\node[black!65, scale=0.45] at (4.000,3.025) {$\diamond$};
\end{scope}
\draw[blue!70!black, solid, line width=0.7pt] (0.45,-0.45) -- (0.93,-0.45);
\node[blue!70!black, scale=0.45] at (0.69,-0.45) {$\bullet$};
\node[font=\scriptsize, anchor=west] at (1.03,-0.45) {WM-Cov};
\draw[red!70!black, dashed, line width=0.7pt] (3.10,-0.45) -- (3.58,-0.45);
\node[red!70!black, scale=0.45] at (3.34,-0.45) {$\circ$};
\node[font=\scriptsize, anchor=west] at (3.68,-0.45) {Risk only};
\draw[green!45!black, dash dot, line width=0.7pt] (5.75,-0.45) -- (6.23,-0.45);
\node[green!45!black, scale=0.45] at (5.99,-0.45) {$\triangle$};
\node[font=\scriptsize, anchor=west] at (6.33,-0.45) {Coverage only};
\draw[black!65, dotted, line width=0.7pt] (8.40,-0.45) -- (8.88,-0.45);
\node[black!65, scale=0.45] at (8.64,-0.45) {$\diamond$};
\node[font=\scriptsize, anchor=west] at (8.98,-0.45) {Random};
\end{tikzpicture}
\caption{Budget curves on executed TeraSim/SUMO event candidates. WM-Cov keeps valid-failure yield close to risk-only selection while maintaining substantially broader coverage and suppressing artifact/invalid selections.}
\label{fig:budget_curves}
\end{figure*}
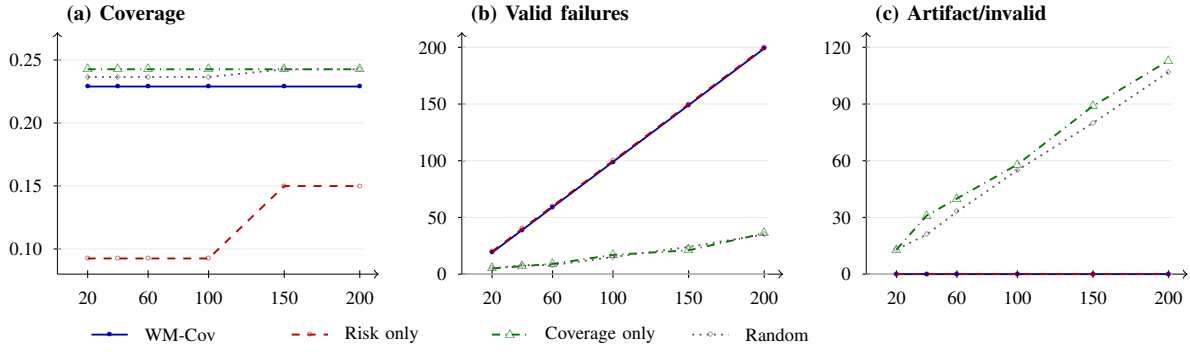

The adequacy-report view makes the difference sharper. Scenario-tag and coverage-only selection both have valid-evidence precision $17/(17+58)=0.23$ on the executed pool, whereas WM-Cov has $99/(99+0)=1.00$ while retaining most of the coverage gain. Thus, the issue is not whether a method can increase a coverage number; it is whether the covered cases remain admissible as safety evidence.
Figure~\ref{fig:budget_curves} visualizes this budgeted selection behavior by comparing how competing policies accumulate valid failures, coverage, and artifact/invalid selections as the rollout budget increases.

The remaining budget and stress-test summaries are reported in Table~\ref{tab:wm_like_budget_100}, Table~\ref{tab:wm_like_sensitivity}, and Table~\ref{tab:artifact_stress}: the WM-like mixed-pool budget comparison, the sensitivity grid, and the artifact-contamination stress test.


\begin{table}[t]
\centering
\caption{Executed TeraSim/SUMO event-level candidate distribution.}
\label{tab:event_distribution}
\begin{tabular}{l r}
\toprule
Category & Count \\
\midrule
valid\_ads\_failure & 230 \\
valid\_safety\_critical & 299 \\
artifact\_failure & 690 \\
\midrule
collision\_pedestrian & 140 \\
collision\_vehicle & 90 \\
hard\_brake & 989 \\
\bottomrule
\end{tabular}
\end{table}
\begin{table}[t]
\centering
\caption{Budget-100 selection trade-off on executed TeraSim/SUMO event candidates.}
\label{tab:budget_100}
\resizebox{\columnwidth}{!}{%
\begin{tabular}{l r r r r r}
\toprule
Method & Coverage & Valid failures & Artifact/invalid & Mean risk & Mean realism \\
\midrule
Random & 0.2364 & 15 & 55 & 0.7759 & 0.6120 \\
Scenario tags & 0.2426 & 17 & 58 & 0.7738 & 0.5998 \\
Risk only & 0.0924 & 100 & 0 & 0.9600 & 0.8600 \\
Realism only & 0.1499 & 100 & 0 & 0.9424 & 0.8600 \\
Coverage only & 0.2426 & 17 & 58 & 0.7738 & 0.5998 \\
WM-Cov & 0.2289 & 99 & 0 & 0.9583 & 0.8598 \\
\bottomrule
\end{tabular}%
}
\end{table}
\begin{table}[t]
\centering
\caption{Budget-100 selection trade-off on WM-like mixed candidate pool.}
\label{tab:wm_like_budget_100}
\resizebox{\columnwidth}{!}{%
\begin{tabular}{l r r r r r r}
\toprule
Method & Coverage & Valid failures & Failure modes & Artifact/invalid & Mean risk & Mean realism \\
\midrule
Random & 0.4674 & 71 & 5 & 7 & 0.8689 & 0.7918 \\
Scenario tags & 0.5144 & 60 & 6 & 8 & 0.8786 & 0.7929 \\
Risk only & 0.4100 & 61 & 6 & 25 & 0.9949 & 0.7053 \\
Realism only & 0.3208 & 70 & 3 & 0 & 0.7795 & 0.9537 \\
Coverage only & 0.5144 & 47 & 6 & 2 & 0.8765 & 0.8165 \\
WM-Cov & 0.4874 & 76 & 6 & 0 & 0.9686 & 0.8932 \\
\bottomrule
\end{tabular}%
}
\end{table}
\begin{table}[t]
\centering
\caption{WM-like candidate sensitivity summary across 16 weight settings.}
\label{tab:wm_like_sensitivity}
\begin{tabular}{l r}
\toprule
Metric & Range \\
\midrule
Coverage & 0.4874--0.4874 \\
Valid failures & 76--77 \\
Artifact/invalid & 0--0 \\
\bottomrule
\end{tabular}
\end{table}
\begin{table}[t]
\centering
\caption{Artifact-contamination stress test on WM-like candidate pools.}
\label{tab:artifact_stress}
\resizebox{\columnwidth}{!}{%
\begin{tabular}{r r r r r r}
\toprule
Artifact ratio & Risk valid & Risk artifact & WM-Cov valid & WM-Cov artifact & WM-Cov cov. \\
\midrule
0.10 & 67.33 & 25.67 & 81.00 & 0.00 & 0.4863 \\
0.25 & 37.33 & 53.67 & 75.67 & 0.00 & 0.4905 \\
0.40 & 29.00 & 59.00 & 64.33 & 0.00 & 0.4917 \\
0.55 & 21.33 & 67.00 & 57.33 & 0.00 & 0.4936 \\
\bottomrule
\end{tabular}%
}
\end{table}

The WM-like mixed trace pool provides a harder adequacy test because it includes valid failures, valid safety-critical cases, duplicate failures, and artifact-like failures in the same source. Table~\ref{tab:wm_like_budget_100} shows that risk-only selection achieves high mean risk, but it admits 25 artifact/invalid cases and selects fewer valid failures than WM-Cov. Scenario-tag and coverage-only selection obtain the largest coverage value, but they sacrifice valid-failure yield or admit artifacts. WM-Cov selects 76 valid failures, covers all six observed failure modes, and admits zero artifact/invalid cases while maintaining high risk and realism.

On this pool, risk-only selection has valid-evidence precision $61/(61+25)=0.71$, while WM-Cov reaches $1.00$. This diagnostic is important for WM-style testing because a world model that produces many high-risk traces can still waste the downstream testing budget if high-risk outputs include duplicates or artifacts.

Table~\ref{tab:wm_like_sensitivity} reports the WM-like sensitivity grid. Across 16 weight settings, WM-Cov maintains zero artifact/invalid selections, a stable coverage value, and 76--77 valid failures. This does not establish global optimality, but it indicates that the proposed adequacy behavior is not a single-weight artifact in the current evaluation.

Table~\ref{tab:artifact_stress} reports the artifact-contamination stress test. As the artifact-like ratio increases from 0.10 to 0.55, risk-only selection increasingly spends the budget on artifact/invalid traces, rising from 25.67 to 67.00 artifact selections on average. WM-Cov maintains zero artifact selections across all tested contamination levels while preserving nontrivial valid-failure yield and stable coverage. This stress test supports the need to evaluate world-model-style testing by audited evidence adequacy rather than by raw danger or risk scores alone.

Together, the two pools show complementary evidence. The executed TeraSim/SUMO pool shows that simulator-generated events can contain many artifact failures, so validity audit is necessary before counting evidence. The WM-like pool illustrates why world-model-style testing should jointly report coverage, validity, failure diversity, risk, and realism under budget. In the 141-row author-verified audit, artifact precision and recall are both 1.00; exact agreement is 0.411 and Cohen's kappa is 0.281, mainly due to duplicate-boundary disagreements.

The DriveArena matrix in Table~\ref{tab:drivearena_closed_loop_batch} and Fig.~\ref{fig:drivearena_formal_matrix} gives the corresponding real provider-chain result. Across 360 requested TrafficManager--WorldDreamer--planner attempts, 304 are fully realized and 56 are partial; none are completely unrealized. This means that every request leaves some auditable artifact, but not every request becomes full planner-conditioned evidence. The heatmap also shows that realization is not a provider constant: VAD realizes all 5.0-s cells, UniAD has a complete 7.5-s default-cloudy realization gap, and both planners show lower realization at the longer horizon. The disjoint route-slice check in Table~\ref{tab:drivearena_second_route_slice} tests whether this requested-to-realized accounting behavior is tied to a single ego-route slice. These differences are exactly why WM-Cov keeps requested attempts, partial realizations, and fully realized traces as separate adequacy objects.

Figure~\ref{fig:interactive_adequacy_curves} reports the stopping-oriented view on the WM-like scenario family. Under a windowed rule that stops when valid-coverage gain and new valid failure modes saturate while artifact rate remains bounded, WM-Cov reaches a stop point at 66 traces with zero artifact/invalid selections and valid-evidence precision of 1.00. Coverage-only selection reaches higher final valid coverage, but it requires 109 traces and admits artifact/invalid evidence. Risk-only selection stops early but covers only a narrow valid region. This illustrates the adequacy trade-off considered here: within a specified scenario family and audit rubric, a campaign should not be judged sufficient merely because it finds failures or covers bins; it should accumulate valid interaction evidence without spending the rollout budget on artifacts or duplicates.

\input{figures/fig5_interactive_adequacy_curve_tikz}

\section{Discussion}
\label{sec:discussion}

The results support an evaluation layer above world models and generative simulators. Failure generation alone is not sufficient for safety testing: open-ended counterfactual generation can expose rare events, but it can also produce duplicates, unrealistic dynamics, map inconsistencies, conditioning failures, and operationally invalid cases. WM-Cov therefore treats a world model as interactive test infrastructure rather than as a test oracle. Evidence is created only after traces are mapped into requested and realized feature spaces, audited, and accumulated by explicit adequacy criteria.

This distinction is the main practical consequence for testing. A world model may be valuable precisely because it can react to the ego planner, but that same adaptivity means that the tester cannot inherit sufficiency arguments from fixed scenario replay. The relevant evidence unit is not a prompt, video, or logged collision; it is a planner-conditioned trace whose requested condition, realized interaction, validity status, and marginal contribution to the remaining coverage gaps are all known. WM-Cov makes those quantities auditable.

The interactive framing is important for interpreting failures. A collision discovered with one ego planner is not necessarily a reusable fixed scenario for every planner. It is evidence that the tested planner failed under a particular scenario family and realization. To evaluate a different planner, the family should be re-instantiated and closed-loop interaction should be regenerated. WM-Cov therefore distinguishes planner-specific failure evidence from potentially transferable safety-critical conditions.

WM-Cov does not replace scenario-based testing taxonomies. It transfers their strongest idea, adequacy through coverage and sufficiency reasoning, into an interactive generative setting where the trace pool is large, adaptive, and partly unreliable. The current evidence should be read in tiers: TeraSim/SUMO events and WM-like stress pools support the accounting logic, the author-verified audit supports explicit validity labels, DriveArena TrafficManager and SUMO/TraCI runs support adapter feasibility over real DriveArena assets, and the 360-request TrafficManager--WorldDreamer--UniAD/VAD provider matrix plus an 80-request disjoint route-slice check support planner-, horizon-, prompt-, and route-slice-conditioned requested-to-realized provider accounting. Larger cross-map and broader planner validation remains the next empirical tier.

\section{Limitations}
\label{sec:limitations}

The current evidence is intentionally tiered. It includes WM-like stress tests, executed generative-simulator events, DriveArena TrafficManager and SUMO/TraCI runs, a real DriveArena TrafficManager--WorldDreamer--UniAD/VAD provider matrix, and a disjoint route-slice robustness check. The DriveArena evidence covers two planners, two horizons, six prompt conditions, and one additional ego-route slice, but it does not validate reference-image conditioning or broad cross-planner generality across maps, traffic densities, and additional planners. The claims should therefore be read as evidence for the evaluation paradigm and accounting logic, not as a benchmark of a particular neural world model.

The taxonomy and validity labels are operational abstractions that should be evaluated across additional maps, simulators, planners, and independently audited cases. The requested-to-realized layer also requires provider-specific instrumentation: if a provider cannot report which prompt, seed, route, and planner produced a trace, then stopping claims become weak. Duplicate evidence likewise depends on the source adapter, especially when one seed contains many actors and event times.

The artifact-contamination stress test is synthetic by design. It isolates whether the selection layer rejects misleading generated evidence as artifact prevalence increases, but it does not characterize the artifact distribution of any deployed neural world model. The DriveArena provider matrix confirms that generated artifacts, planner responses, partial realizations, and condition-specific provider failures can be ingested and accounted for, but it remains too narrow to estimate artifact distributions across maps, prompts, weather, and planners. Future work should estimate these distributions from larger generated rollouts and test planner transferability by re-instantiating the same scenario family with multiple ego planners.

\section{Conclusion}
\label{sec:conclusion}

World-model-style simulation changes autonomous-driving testing from fixed scenario replay to interactive scenario-family evaluation. The adequacy question is therefore not how many visually plausible or dangerous rollouts a generator can produce, but whether a stated testing intent has accumulated enough requested, realized, valid, and non-redundant evidence for the ego planner under test. WM-Cov turns this question into an auditable measurement contract: a provider-agnostic adapter records each trace, the requested--realized--valid ledger separates test intent from generated evidence, the audit layer removes artifacts and duplicates from sufficiency accounting, and the selection layer reports how evidence accumulates under a finite testing budget.

The experiments support this contract in three tiers. Executed TeraSim/SUMO events and WM-like stress pools show that generated danger is not automatically useful test evidence; the author-verified audit shows that validity categories can be inspected; and the 360-request DriveArena TrafficManager--WorldDreamer matrix plus an 80-request disjoint route-slice check move the same accounting logic into a real closed-loop provider chain. Their main lesson is that even when every request produces some artifact, the number of fully realized traces depends on planner, horizon, prompt, and route slice. The present evidence does not set a universal run count or establish safety sufficiency. It shows what must be measured before such claims can be made: requested-to-realized coverage, validity, duplicate control, partial-realization accounting, and stopping-oriented evidence reports.

\section*{Data and Code Availability}

The public replication package is available at
\url{https://github.com/cuijianxun/wm-cov}. It includes WM-Cov scripts, adapter
schemas, derived evidence tables, aggregate ledgers, and figure assets, but not
license-restricted third-party datasets, model checkpoints, or raw provider
frames.

\section*{Acknowledgment}

This research is funded by the Chongqing Natural Science Foundation Innovation
and Development Joint Fund (Changan Automobile) (Grant No.
CSTB2024NSCQ-LZX0157).

\bibliographystyle{IEEEtran}
\bibliography{references}

@article{guan2025world,
  title = {World Models for Autonomous Driving: An Initial Survey},
  author = {Guan, Yanchen and Liao, Haicheng and Li, Zhenning and Hu, Jia and Yuan, Runze and Zhang, Guohui and Xu, Chengzhong},
  journal = {IEEE Transactions on Intelligent Vehicles},
  pages = {1--17},
  year = {2025},
  doi = {10.1109/TIV.2024.3398357},
  url = {https://doi.org/10.1109/TIV.2024.3398357}
}

@article{tang2023survey,
  title = {A Survey on Automated Driving System Testing: Landscapes and Trends},
  author = {Tang, Shuncheng and Zhang, Zhenya and Zhang, Yi and Zhou, Jixiang and Guo, Yan and Liu, Shuang and Guo, Shengjian and Li, Yan-Fu and Ma, Lei and Xue, Yinxing and Liu, Yang},
  journal = {ACM Transactions on Software Engineering and Methodology},
  volume = {32},
  number = {5},
  pages = {1--62},
  year = {2023},
  month = jul,
  doi = {10.1145/3579642},
  url = {https://doi.org/10.1145/3579642}
}

@article{laurent2023parameter,
  title = {Parameter Coverage for Testing of Autonomous Driving Systems under Uncertainty},
  author = {Laurent, Thomas and Klikovits, Stefan and Arcaini, Paolo and Ishikawa, Fuyuki and Ventresque, Anthony},
  journal = {ACM Transactions on Software Engineering and Methodology},
  volume = {32},
  number = {3},
  pages = {1--31},
  year = {2023},
  month = apr,
  doi = {10.1145/3550270},
  url = {https://doi.org/10.1145/3550270}
}

@inproceedings{neelofar2024towards,
  title = {Towards Reliable AI: Adequacy Metrics for Ensuring the Quality of System-Level Testing of Autonomous Vehicles},
  author = {Neelofar and Aleti, Aldeida},
  booktitle = {Proceedings of the IEEE/ACM 46th International Conference on Software Engineering},
  series = {ICSE '24},
  pages = {1--12},
  year = {2024},
  publisher = {ACM},
  doi = {10.1145/3597503.3623314},
  url = {https://doi.org/10.1145/3597503.3623314}
}

@inproceedings{degelder2024coverage,
  title = {Coverage Metrics for a Scenario Database for the Scenario-Based Assessment of Automated Driving Systems},
  author = {de Gelder, Erwin and Buermann, Maren and Op Den Camp, Olaf},
  booktitle = {2024 IEEE International Automated Vehicle Validation Conference},
  pages = {1--8},
  year = {2024},
  publisher = {IEEE},
  doi = {10.1109/IAVVC63304.2024.10786405},
  url = {https://doi.org/10.1109/IAVVC63304.2024.10786405}
}

@article{degeldersingh2025scenario,
  title = {Scenario Metrics for the Safety Assurance Framework of Automated Vehicles: A Review of Its Application},
  author = {de Gelder, Erwin and Singh, Tajinder and Hadj-Selem, Fouad and Vidal Bazan, Sergi and Op den Camp, Olaf},
  journal = {Vehicles},
  volume = {7},
  number = {3},
  pages = {100},
  year = {2025},
  doi = {10.3390/vehicles7030100},
  url = {https://doi.org/10.3390/vehicles7030100}
}

@article{chodowiec2026odd,
  title = {{ODD} and Behavior-Based Approach to Scenario Coverage for Automated Driving Systems Testing},
  author = {Chodowiec, Emil and Irvine, Patrick and Tiele, Justin-Kiyoshi and Takenaka, Kazuhito and Zhang, Xizhe and Khastgir, Siddartha and Jennings, Paul A.},
  journal = {IEEE Access},
  volume = {14},
  pages = {33117--33139},
  year = {2026},
  doi = {10.1109/ACCESS.2026.3665396},
  url = {https://doi.org/10.1109/ACCESS.2026.3665396}
}

@article{zhu2025triggering,
  title = {Leveraging Triggering Conditions for Efficient Scenario-Based Testing of Automated Vehicles},
  author = {Zhu, Zhijing and Philipp, Robin and Howar, Falk},
  journal = {SAE International Journal of Connected and Automated Vehicles},
  volume = {8},
  number = {4},
  pages = {537--549},
  year = {2025},
  doi = {10.4271/12-08-04-0035},
  url = {https://doi.org/10.4271/12-08-04-0035}
}

@article{ding2023survey,
  title = {A Survey on Safety-Critical Driving Scenario Generation---A Methodological Perspective},
  author = {Ding, Wenhao and Xu, Chejian and Arief, Mansur and Lin, Haohong and Li, Bo and Zhao, Ding},
  journal = {IEEE Transactions on Intelligent Transportation Systems},
  volume = {24},
  number = {7},
  pages = {6971--6988},
  year = {2023},
  month = jul,
  doi = {10.1109/TITS.2023.3259322},
  url = {https://doi.org/10.1109/TITS.2023.3259322}
}

@inproceedings{andrews2015active,
  title = {Active World Model for Testing Autonomous Systems Using {CEFSM}},
  author = {Andrews, Anneliese and Abdelgawad, Mahmoud and Gario, Ahmed},
  booktitle = {Proceedings of the 12th Workshop on Model-Driven Engineering, Verification and Validation},
  series = {CEUR Workshop Proceedings},
  volume = {1514},
  pages = {1--10},
  year = {2015},
  url = {https://ceur-ws.org/Vol-1514/paper1.pdf}
}

@inproceedings{andrews2016world,
  title = {World Model for Testing Autonomous Systems Using Petri Nets},
  author = {Andrews, Anneliese and Abdelgawad, Mahmoud and Gario, Ahmed},
  booktitle = {2016 IEEE 17th International Symposium on High Assurance Systems Engineering},
  pages = {65--69},
  year = {2016},
  publisher = {IEEE},
  doi = {10.1109/HASE.2016.11},
  url = {https://doi.org/10.1109/HASE.2016.11}
}

@misc{hu2023gaia,
  title = {GAIA-1: A Generative World Model for Autonomous Driving},
  author = {Hu, Anthony and Russell, Lloyd and Yeo, Hudson and Murez, Zak and Fedoseev, George and Kendall, Alex and Shotton, Jamie and Corrado, Gianluca},
  year = {2023},
  eprint = {2309.17080},
  archivePrefix = {arXiv},
  primaryClass = {cs.CV},
  url = {https://arxiv.org/abs/2309.17080},
  note = {arXiv preprint}
}

@inproceedings{yang2025drivearena,
  title = {DriveArena: A Closed-Loop Generative Simulation Platform for Autonomous Driving},
  author = {Yang, Xuemeng and Wen, Licheng and Wei, Tiantian and Ma, Yukai and Mei, Jianbiao and Li, Xin and Lei, Wenjie and Fu, Daocheng and Cai, Pinlong and Dou, Min and He, Liang and Liu, Yong and Shi, Botian and Qiao, Yu},
  booktitle = {2025 IEEE/CVF International Conference on Computer Vision},
  pages = {26933--26943},
  year = {2025},
  publisher = {IEEE},
  doi = {10.1109/ICCV51701.2025.02500},
  url = {https://doi.org/10.1109/ICCV51701.2025.02500}
}

@misc{yang2026resim,
  title = {ReSim: Reliable World Simulation for Autonomous Driving},
  author = {Yang, Jiazhi and Chitta, Kashyap and Gao, Shenyuan and Chen, Long and Shao, Yuqian and Jia, Xiaosong and Li, Hongyang and Geiger, Andreas and Yue, Xiangyu and Chen, Li},
  year = {2026},
  eprint = {2506.09981},
  archivePrefix = {arXiv},
  primaryClass = {cs.CV},
  version = {v2},
  url = {https://arxiv.org/abs/2506.09981},
  note = {arXiv preprint}
}

@misc{sun2025terasim,
  title = {TeraSim: Uncovering Unknown Unsafe Events for Autonomous Vehicles through Generative Simulation},
  author = {Sun, Haowei and Yan, Xintao and Qiao, Zhijie and Zhu, Haojie and Sun, Yihao and Wang, Jiawei and Shen, Shengyin and Hogue, Darian and Ananta, Rajanikant and Johnson, Derek and Stevens, Greg and McGuire, Greg and Wei, Yifan and Zheng, Wei and Sun, Yong and Fukai, Yasuo and Liu, Henry X.},
  year = {2025},
  eprint = {2503.03629},
  archivePrefix = {arXiv},
  primaryClass = {cs.RO},
  version = {v4},
  url = {https://arxiv.org/abs/2503.03629},
  note = {arXiv preprint}
}

@article{gao2026foundation,
  title = {Foundation Models in Autonomous Driving: A Survey on Scenario Generation and Scenario Analysis},
  author = {Gao, Yuan and Piccinini, Mattia and Zhang, Yuchen and Wang, Dingrui and Moller, Korbinian and Brusnicki, Roberto and Zarrouki, Baha and Gambi, Alessio and Totz, Jan Frederik and Storms, Kai and Peters, Steven and Stocco, Andrea and Alrifaee, Bassam and Pavone, Marco and Betz, Johannes},
  journal = {IEEE Open Journal of Intelligent Transportation Systems},
  pages = {1--1},
  year = {2026},
  doi = {10.1109/OJITS.2026.3660686},
  url = {https://doi.org/10.1109/OJITS.2026.3660686}
}

@article{song2026generative,
  title = {Generative AI for Testing of Autonomous Driving Systems: A Survey},
  author = {Song, Qunying and Ye, He and Harman, Mark and Sarro, Federica},
  journal = {ACM Transactions on Software Engineering and Methodology},
  year = {2026},
  doi = {10.1145/3806653},
  url = {https://doi.org/10.1145/3806653}
}

@article{tuncali2025synthetic,
  title = {Synthetic versus Real: An Analysis of Critical Scenarios for Autonomous Vehicle Testing},
  author = {Tuncali, Cumhur Erkan and Yaghoubi, Shakiba and Fainekos, Georgios and Ito, Hisahiro},
  journal = {Automated Software Engineering},
  volume = {32},
  number = {2},
  year = {2025},
  doi = {10.1007/s10515-025-00499-4},
  url = {https://doi.org/10.1007/s10515-025-00499-4}
}

\end{document}